\documentclass[preprint,3p]{elsarticle}
\usepackage{epstopdf}
\usepackage{caption}
\usepackage{setspace}
\usepackage{amssymb}
\usepackage{multirow}
\usepackage{amsthm}
\usepackage{booktabs}
\usepackage{amsmath}
\usepackage{amssymb}
\usepackage{mathrsfs}
\usepackage{longtable}
\usepackage{lscape}
\usepackage{url}
\usepackage{tabularx}
\usepackage{epsfig}
\usepackage{colortbl}
\usepackage{subfigure}
\usepackage{tikz,mathpazo}
\usepackage{graphicx}
\usepackage{natbib}
\usepackage[noend]{algpseudocode}
\usepackage{algorithm}
\usepackage{algorithmicx}

\newtheorem{myDef}{Definition}

\usetikzlibrary{shapes.geometric, arrows}
\biboptions{numbers,sort&compress}

\begin{document}
\begin{frontmatter}
\title{Ordinal relative belief entropy}
\author[address1]{Yuanpeng He}
\author[address1]{Yong Deng \corref{label1}}

\address[address1]{Institute of Fundamental and Frontier Science, University of Electronic Science and Technology of China, Chengdu, China}
\cortext[label1]{Corresponding author: Yong Deng, Institute of Fundamental and Frontier Science, University of Electronic Science and Technology of China, Chengdu, 610054 Sichuan, China. Email address:  dengentropy@uestc.edu.cn and ydeng@swu.edu.cn.}
\begin{abstract}
Specially customised Entropies are widely applied in measuring the degree of uncertainties existing in the frame of discernment. However, all of these entropies regard the frame as a whole that has already been determined which dose not conform to actual situations. In real life, everything comes in an order, so how to measure uncertainties of the dynamic process of determining sequence of propositions contained in a frame of discernment is still an open issue and no related research has been proceeded. Therefore, a novel ordinal entropy to measure uncertainties of the frame of discernment considering the order of confirmation of propositions is proposed in this paper. Compared with traditional entropies, it manifests effects on degree of uncertainty brought by orders of propositions existing in a frame of discernment. Besides, some numerical examples are provided to verify the correctness and validity of the proposed entropy in this paper.
\end{abstract}
\end{frontmatter}
\section{Introduction}
How to properly handle uncertainty of given information has been a hot topic. A variety of theories have been proposed to extract useful information from uncertainty, such as $D$ numbers \cite{Xiao2019a, LiubyDFMEA, Deng2019}, fuzzy set theory \cite{Xue2020entailmentIJIS, Zadeh1965, 6608375, 8944285}, $Z$ number \cite{Liu2019b} and so on \cite{Yager2014, Pan2020, Yager2014a}. On the basis of proposed theories, lots of corresponding applications have been made, like target recognition \cite{Pan2020}, pattern classification \cite{Liu2019, Song2018, articledd}, decision making \cite{Song2019a, Yager2009, Han2019, fei2019mcdm} and so on \cite{Fei2019, articlesdsdsd,Li2020generateTDBF}.

Before processing uncertain information, it is necessary to measure level of uncertainty of given information. One of the efficient way is utilizing entropy, the most representative work is Shannon entropy \cite{inbook}, which reflects a volume of information, namely amount of uncertainty contained in provided information. Although some related entropies also have been proposed, such as interval-valued entropy \cite{articleYager}, motion entropy \cite{articleChakraborty} and non-additive entropy \cite{articleTsallis}, the Shannon entropy is still regarded as the most effective tool to judge conditions of given information. In order to describe every incident systematically, an efficient theory, Dempster-Shafer evidence theory (D-S evidence theory) \cite{book, Dempster1967Upper}, is developed to offer a more adaptable solution to handle relations and conflicts among propositions existing in frames of discernment. Utilizing Dempster-Shafer evidence theory, as a generalization of Shannon entropy, Deng entropy \cite{Deng2020ScienceChina} is proposed to measure conditions of uncertainty of frames of discernment, which considers constitution of each proposition and build a bridge on measuring method between informatics and D-S evidence theory.

However, all of the previous work ignore a crucial factor which is the order of confirmation of propositions. What should be pointed out is that an ascertainment of a proposition has a effect on values of other propositions contained in the same frame of discernment. Some worthy work has considered the difference between sets and orderly sets \cite{articleSunberg, articleCheng} which firstly notice the influences of orders of propositions. Therefore, it is necessary to take sequences of propositions into consideration in D-S evidence theory. Nevertheless, how to measure degrees of uncertainty of an ordinal frame of discernment is still an open issue. To address this problem, an ordinal relative belief entropy is proposed, which considers effects brought by sequences of propositions in frames of discernment. Compared with previously proposed entropies, it indicates fluctuations of uncertainty of the whole system of discernment.

This paper is organized as follows. Some related concepts are briefly introduced in the section of preliminaries. In the next section, the definition of proposed entropy is explained in details. Some numerical examples are offered to verify correctness and validity of the proposed entropy. In the last, the whole passage is summarized in the section of conclusion.

\section{Preliminaries}

In this section, some related concepts are briefly introduced. Lots of relative meaningful researches have been made, such as generalized D-S structures \cite{yager2019generalized, yager1987dempster} and fuzzy linguistic approach \cite{Liu2020a}.

\subsection{Dempster-Shafer evidence theory}
Dempster-Shafer evidences theory (D-S theory) \cite{book, Dempster1967Upper} has been widely accepted in handling multi-source information. Due to its prominent effectiveness, the theory has been applied in many fields, such as complex mass function \cite{Xiao2019b, Xiao2020b}, Bayesian network \cite{Jiang2019Znetwork}, negation theory \cite{Zhang2020a, Li2020, Luo2019negation} and so on \cite{Luo2020vectorIJIS, Gao2019b, Deng2020InformationVolume}.

\begin{myDef}(Frame of discernment)\end{myDef}
Let $\Theta$ be a non-empty and mutually exclusive set which is consist of $n$ random values. Then, set $\Theta$ is named as frame of discernment which is defined as \cite{book, Dempster1967Upper}:
\begin{equation}
	\Theta = \{P_{1}, P_{2}, ..., P_{i}, ..., P_{n}\}
\end{equation}

And the power set of $\Theta$ composed of $2^{n}$ distinctive elements is dented as $2^{\Theta}$ defined as:
\begin{equation}
2^{\Theta} = \{\emptyset, \{P_{1}\}, \{P_{2}\}, ..., \{P_{1}, P_{2}\}, ..., \{P_{1}, P{2}, ..., P_{n}\}, ..., \Theta\}
\end{equation}

\begin{myDef}(Mass function)\end{myDef}
On the base of the definition of frame of discernment, a mapping $m$ is also a mass function from $2^{\Theta}$ to $[0, 1]$ which is defined as \cite{book, Dempster1967Upper}:
\begin{equation}
	m: 2^{\Theta} \rightarrow [0, 1]
\end{equation}

Besides, the mass function also satisfies:
\begin{equation}
	\sum_{A \in 2^{\Theta}}m(A) = 1
\end{equation}
\begin{equation}
	m(\emptyset) = 0
\end{equation}

The mass function is also named as basic probability assignment (BPA) according to D-S evidence theory. If $A$ is a proposition existing in the frame of discernment, then the value of proposition $A$ indicates the support provided by evidences to proposition $A$. Besides, if $m(A) > 0$, A is also called as a focal element. 

Due to effectiveness of D-S evidence theory in handling uncertainties, the theory is also extended to the field of quantum \cite{articlequantum} which alleviates uncertainties at a completely new level.

\subsection{Some entropy theories}
To figure out what is truly contained in uncertainties, it is necessary to find effective ways to measure the degree of uncertainties. Driven by this kind of motivation, entropy theory has been widely utilized to offer a solution to solve this problem \cite{ pan2020Probability, xiao2019multi}. 

\begin{myDef}(Dubois $\&$ Prade’s weighted Hartley entropy)\end{myDef}
When a frame of discernment is given, Dubois $\&$ Prade’s weighted Hartley entropy is defined as \cite{articlefuzzy}:
\begin{equation}
	E_{dp}(m) = \sum_{A \in 2^{\Theta}} m(A)log_{2}|A|
\end{equation}

Where m represents a mass function developed on the basis of the definition of a frame of discernment and $A$ is a proposition existing in the frame. Besides, the value of $m(A)$ denotes the level of support to the proposition $A$ and $|A|$ indicates the cardinality of the proposition.

\begin{myDef}(Deng entropy)\end{myDef}
When a frame of discernment is given, Deng entropy is defined as \cite{Deng2020ScienceChina}:
\begin{equation}
	E_{d}(m) = -\sum_{A \in 2^{\Theta}} m(A)log_{2}\frac{m(A)}{2^{|A|} - 1}
\end{equation}

Where m is a mass function according to the definition of frame of discernment and $A$ is a focal element. The cardinality of proposition $A$ is denoted as $|A|$. Besides, the mass of $m(A)$ represents the support offered by evidences to proposition $A$.

\section{Proposed ordinal relative belief entropy}
However, entropies have been widely utilized to measure uncertainty of a system \cite{yanhy2020entropy, Abellan2017, WEI20114273, zhang2020extension}, but none of them take orders of propositions into consideration. The proposed entropy regards the orders of proposition as a crucial factor in determining the degree of uncertainty to present actual conditions, because occurrences of things are involved with their sequences.
\begin{myDef}(Ordinal frame of discernment)\end{myDef}
All of the elements contained in the ordinal frame of discernment are associated with an order. For a ordinal set $\Theta_{order} = \{P_{1}^{1}, P_{2}^{2}, ..., P_{n}^{n}\}$, the subscript represents the number of propositions existing in the frame of discernment and the superscript denotes the sequence of propositions. Then, the properties all the elements satisfy are listed as follows:
\begin{itemize}
	\item The proposition $P_{1}^{1}$ is supposed to be ascertained before $P_{2}^{2}$ and the rest of propositions can be confirmed in the same manner.
	\item All the propositions existing in the frame of discernment only have the relations mentioned in the last point. Except for that, the definitions of the propositions remains consistent with the ones defined in the classic D-S evidence theory.
	\item In the process of determining every proposition, uncertainty of the whole system is further confirmed.
\end{itemize}

\textbf{Note: }Subscripts are not supposed to be the same as superscripts. In order to simplify the process of calculation, a constraint that subscript is equal to superscript is made.

\textbf{Case 1: }Assume three propositions $P_{1}^{1}, P_{2}^{2}, P_{3}^{3}$ are contained in an ordinal frame of discernment which can be presented as $\Theta_{order} = \{\{P_{1}^{1}\}, \{P_{2}^{2}\}, \{P_{3}^{3}\}\}$. Compared to classic D-S evidence theory, the difference is that the proposition $P_{1}^{1}$ must come first, then $P_{2}^{2}$ is confirmed and $P_{3}^{3}$ is ascertained in the last. Proposition $P_{2}^{2}$ or $P_{3}^{3}$ is not allowed to take the first place. Similarly, Proposition $P_{1}^{1}$ is also not expected to take the place of $P_{2}^{2}$ or $P_{3}^{3}$. If the order is disorganized, then the whole system may obtain a completely new status.
\begin{myDef}Distributed weights for proposed entropy\end{myDef}
Because the frame of discernment is ordinal, the uncertainty of systems is further confirmed with every step of ascertaining a proposition. It is necessary to regard the degrees of importance of determining the whole system of propositions are different. Therefore, assume the number of propositions existing in the ordinal frame of discernment is $n$ and the sequence of a proposition is represented by $m$. Then the process of obtaining the modified values is defined as:

\begin{enumerate} [(1)]
	\item Weights of propositions is $n - m + 1$, the calculation is expressed as:
	\begin{equation}
		Weight_{P_{i}^{j}} = n - m_{P_{i}^{j}} +1
	\end{equation}
	 
	\item Original mass of propositions is denoted as $Mass_{P_{i}^{j}}$. Contemporary intermediate values of propositions are expressed as:
	\begin{equation} Value_{P_{i}^{j}} = Mass_{P_{i}^{i}} \times Weight_{P_{i}^{j}}\end{equation}
	
	\item The step of normalization of intermediate values is expressed as:
	\begin{equation} Value_{P_{i}^{j}}^{Final} = \frac{Value_{P_{i}^{j}}}{\sum_{i = 1 = j}^{n}Value_{P_{i}^{j}}}\end{equation}
\end{enumerate}
\begin{myDef}(Relative belief entropy)\end{myDef}

It is impractical to utilize previously proposed entropy to measure the level of uncertainty of ordinal frames of discernment. Because all of the entropies consider every proposition as an independent unit, which is acceptable in classic frame of discernment. However, determination of every proposition in ordinal frame of discernment affects conditions and values of propositions which are expected to be ascertained in the following. According to conditions described, it is necessary to manifest relations among propositions in the expression of proposed entropy instead of considering each of them solely and ignoring effects single individual has on the whole. Therefore, a relative belief entropy is proposed to solve the problem mentioned above and it is expressed as:

\begin{equation}
 U(P_{i}^{j},P_{a}^{b}) = Value_{P_{i}^{j}}^{Final} \times ln^{((\frac{Value_{P_{i}^{j}}^{Final}}{2^{|P_{i}^{j}|} - 1}) / (\frac{Value_{P_{a}^{b}}^{Final}}{2^{|P_{a}^{b}|} - 1}) + e)}
\end{equation}

The denominators of each minimum division expression is dependent on the number of elements contained in propositions, which is also called the cardinality of a proposition. For example, proposition $\{P_{1}^{1}, P_{2}^{2}, P_{3}^{3}\}$ has three elements, so the value of corresponding denominator is $2^{|P_{1}^{1}, P_{2}^{2}, P_{3}^{3}|} - 1 = 2^{3} - 1 = 7$. 

\textbf{Case 2: }Assume there are three propositions $P_{1}^{1}, P_{2}^{2}, P_{3}^{3}$ and the ordinal frame of discernment is defined as $\{\{P_{1}^{1}\}, \{P_{2}^{2}\}, \{P_{3}^{3}\}\}$. Besides, their values after modification are listed as: $Value_{P_{1}^{1}}^{Final}= \frac{1}{3}, Value_{P_{2}^{2}}^{Final}= \frac{1}{3}, Value_{P_{3}^{3}}^{Final}= \frac{1}{3}$. Then, with respect to proposition $P_{1}^{1}$, the mass of relative belief entropy can expressed as:

$U(P_{1}^{1},P_{2}^{2}) = Value_{P_{1}^{1}}^{Final} \times ln^{((\frac{Value_{P_{1}^{1}}^{Final}}{2^{|P_{1}^{1}|} - 1}) / (\frac{Value_{P_{2}^{2}}^{Final}}{2^{|P_{2}^{2}|} - 1}) + e)} = \frac{1}{3}$,

$U(P_{1}^{1},P_{3}^{3}) = Value_{P_{1}^{1}}^{Final} \times ln^{((\frac{Value_{P_{1}^{1}}^{Final}}{2^{|P_{1}^{1}|} - 1}) / (\frac{Value_{P_{3}^{3}}^{Final}}{2^{|P_{3}^{3}|} - 1}) + e)} = \frac{1}{3}$.
\begin{myDef}(Individual ordinal relative belief entropy)\end{myDef}
In the ordinal frame of discernment, an order exists among propositions, so the calculation of a value of relative belief entropy of a certain proposition is unidirectional. For instance, for proposition $P_{1}^{1}$, the mass about $P_{1}^{1} \rightarrow P_{2}^{2}$ and $P_{1}^{1} \rightarrow P_{3}^{3}$ can be calculated. Besides, with respect to proposition $P_{3}^{3}$, only the mass about $P_{2}^{2} \rightarrow P_{3}^{3}$ can be calculated. The calculation of individual ordinal relative belief entropy is defined as:
\begin{equation}
IU(P_{i}^{j},P_{a}^{b}) = \sum_{b = j+ 1}^{n} U(P_{i}^{j},P_{a}^{b})
\end{equation}

Where n is the sequence of the last proposition. The individual ordinal relative belief entropy can be also considered as a staged entropy, because only one proposition is confirmed at this stage and conditions of other propositions are not taken into account. An example is provided to offered the process of calculation. 

\textbf{Note: }As for the last proposition, the value of entropy of it is regarded as 0. Because when all the preceding propositions is determined, the whole system of frame of discernment is also determined. In other words, relations exist among propositions and an ascertainment of a proposition affects situations of other propositions. 

\textbf{Case 3: }Assume there are three propositions $P_{1}^{1}, P_{2}^{2}, P_{3}^{3}$ and the ordinal frame of discernment is defined as $\{\{P_{1}^{1}\}, \{P_{2}^{2}\}, \{P_{3}^{3}\}\}$. Besides, their values after modification are listed as: $Value_{P_{1}^{1}}^{Final}= \frac{1}{3}, Value_{P_{2}^{2}}^{Final}= \frac{1}{3}, Value_{P_{3}^{3}}^{Final}= \frac{1}{3}$. And the process of calculation of values of propositions is expressed as:

$IU(P_{1}^{1},P_{a}^{b}) = \sum_{b = j+ 1}^{n} U(P_{1}^{1},P_{a}^{b}) = Value_{P_{1}^{1}}^{Final} \times ln^{((\frac{Value_{P_{1}^{1}}^{Final}}{2^{|P_{1}^{1}|} - 1}) / (\frac{Value_{P_{2}^{2}}^{Final}}{2^{|P_{2}^{2}|} - 1}) + e)} + Value_{P_{1}^{1}}^{Final} \times ln^{((\frac{Value_{P_{1}^{1}}^{Final}}{2^{|P_{1}^{1}|} - 1}) / (\frac{Value_{P_{3}^{3}}^{Final}}{2^{|P_{3}^{3}|} - 1}) + e)} \\ = \frac{1}{3} + \frac{1}{3} = \frac{2}{3}$

$IU(P_{2}^{2},P_{a}^{b}) = \sum_{b = j+ 1}^{n} U(P_{2}^{2},P_{a}^{b}) = Value_{P_{2}^{2}}^{Final} \times ln^{((\frac{Value_{P_{2}^{2}}^{Final}}{2^{|P_{2}^{2}|} - 1}) / (\frac{Value_{P_{3}^{3}}^{Final}}{2^{|P_{3}^{3}|} - 1}) + e)} = \frac{1}{3}$

$IU(P_{3}^{3},P_{a}^{b}) = 0$

Because the frame of discernment is ordinal, the process of calculation is unidirectional which also conforms to the definition of ordinal frame of discernment. Every value of ordinal relative belief entropy manifest uncertainty at the stage of confirming current propositions. With increase of the number of propositions confirmed, degree of uncertainty of the whole system is further confirmed which is presented by the values of ordinal relative belief entropy. The proposed entropy properly measures conditions of every component in the whole system.
\begin{myDef}(Integral ordinal relative belief entropy)\end{myDef}
The integral ordinal relative belief entropy is a synthesis of individual ordinal relative belief entropy which measures conditions of the whole system. In other words, integral ordinal relative belief entropy considers every stage of the complete frame. The process of calculation is defined as:
\begin{equation}
INU(P_{i}^{j},P_{a}^{b}) = \sum_{j = 1}^{n} IU(P_{i}^{j},P_{a}^{b})
\end{equation}

\textbf{Case 4: }Assume there are three propositions $P_{1}^{1}, P_{2}^{2}, P_{3}^{3}$ and the ordinal frame of discernment is defined as $\{\{P_{1}^{1}\}, \{P_{2}^{2}\}, \{P_{3}^{3}\}\}$. Besides, their values after modification are listed as: $Value_{P_{1}^{1}}^{Final}= \frac{1}{3}, Value_{P_{2}^{2}}^{Final}= \frac{1}{3}, Value_{P_{3}^{3}}^{Final}= \frac{1}{3}$. And the process of calculation of values of propositions is expressed as:

$INU(P_{i}^{j},P_{a}^{b}) = \sum_{j = 1}^{n} IU(P_{i}^{j},P_{a}^{b}) = IU(P_{1}^{1},P_{a}^{b}) + IU(P_{2}^{2},P_{a}^{b}) + IU(P_{3}^{3},P_{a}^{b}) = 1$

Then, this is a final description of level of uncertainty about the system.

\subsection{Measurement about classic frame of discernment using ordinal relative belief entropy}
In the last section, a new entropy is proposed to measure degree of uncertainty of ordinal frame of discernment \cite{ deng2021fuzzymembershipfunction} which manifests an important role order plays in influencing degrees of uncertainty of the whole system \cite{article5555}. However, the proposed entropy can be also utilized to describe conditions about a classic frame of discernment considering every kind of order of a certain frame. For example, if there are three propositions existing in a frame of discernment, then every place corresponding to an order can be held by each of propositions. Therefore, it is necessary to find out all kinds of combinations and get an average utilizing sums of values from different combinations. Therefore, a measurement of classic frame of discernment is made. The process of calculation is defined as:

\begin{enumerate} [(1)]
\item Find out all kinds of the possible combinations.
\item Get a sum of values from combinations using definition of ordinal relative belief entropy.
\item Calculate an average of the sum according to the number of kinds of combinations to get a final mass.
\end{enumerate}

\textbf{Case 5: }Assume there are three propositions $P_{1}, P_{2}, P_{3}$ and the frame of discernment is defined as $\{\{P_{1}\}, \{P_{2}\}, \{P_{3}\}\}$. Besides, their values are listed as: $P_{1}= \frac{1}{6}, P_{2}= \frac{1}{2}, P_{3}= \frac{1}{3}$. In this case, in order to conform to usage of previously proposed entropy, the superscript is concealed to focus on effects brought by sequences. All the possible combinations are listed below:

$\{P_{1}, P_{2}, P_{3}\}$,  $\{P_{1}, P_{3}, P_{2}\}$,  $\{P_{2}, P_{1}, P_{3}\}$,  $\{P_{2}, P_{3}, P_{1}\}$, $\{P_{3}, P_{1}, P_{2}\}$, $\{P_{3}, P_{2}, P_{1}\}$.

In all, there are six different combinations, which also means there are six different ordinal frame of discernment. Using ordinal relative belief entropy, their values calculated are listed below:

$INU_{Frame_{1}}(P_{i}^{j},P_{a}^{b}) = 1.6624$,  $INU_{Frame_{2}}(P_{i}^{j},P_{a}^{b}) = 1.3267$, $INU_{Frame_{3}}(P_{i}^{j},P_{a}^{b}) = 2.9388$, 

$INU_{Frame_{4}}(P_{i}^{j},P_{a}^{b}) = 3.1569$, $INU_{Frame_{5}}(P_{i}^{j},P_{a}^{b}) = 2.0190$, $INU_{Frame_{6}}(P_{i}^{j},P_{a}^{b}) = 2.6049$.

\textbf{Note: }All of the superscript is concealed to be consistent to environment of application.

Because the number of combinations is six, the final description about the unordered frame of discernment is the sum of six values of $INU$ divides six.

$INU_{Final} = \frac{INU_{Frame_{1}}(P_{i}^{j},P_{a}^{b})+ INU_{Frame_{2}}(P_{i}^{j},P_{a}^{b}) + INU_{Frame_{3}}(P_{i}^{j},P_{a}^{b}) + INU_{Frame_{4}}(P_{i}^{j},P_{a}^{b}) + INU_{Frame_{5}}(P_{i}^{j},P_{a}^{b}) + INU_{Frame_{6}}(P_{i}^{j},P_{a}^{b})}{6} \\= 2.1181$

Utilizing two entropies mentioned above, two degrees of uncertainty about frame of discernment provided are calculated. The results are listed in Table \ref{1}.

\begin{table}[htbp]	
	\centering
	\caption{Results of three kinds of entropies}
	\begin{tabular}{ccc}
		\toprule  
		Dubois $\&$ Prade’s weighted Hartley entropy&Deng entropy&Proposed entropy \\ 
		\midrule  
		0&1.0113&2.1181 \\
		\bottomrule  
		\label{1}
	\end{tabular}
\end{table}

In this case, the effectiveness of proposed ordinal entropy in measuring degrees of uncertainty of classic frame of discernment is verified, which considers every kind of order to obtain a proper description about unordered frame of discernment.

\section{Numerical examples}
In this section, some examples are provided to verify effectiveness of proposed entropy in illustrating how an order of a ordinal frame of discernment affects levels of uncertainty of whole systems. 

\textbf{Example 1: }Assume there are three propositions $P_{1}^{1}, P_{2}^{2}, P_{3}^{3}$ and the ordinal frame of discernment is defined as $\{\{P_{1}\}, \{P_{2}\}, \{P_{3}\}\}$. Besides, their values after modification are listed as: $Mass_{P_{1}^{1}}= \frac{1}{3}, Mass_{P_{2}^{2}}= \frac{1}{3}, Mass_{P_{3}^{3}}= \frac{1}{3}$. The process of calculation with respect to the sequence $\{P_{1}^{1}, P_{2}^{2}, P_{3}^{3}\}$ is given as:

First, obtain weights corresponding to propositions.

$Weight_{P_{1}^{1}} = n - m_{P_{1}^{1}} +1 = 3$, $Weight_{P_{2}^{2}} =2$, $Weight_{P_{3}^{3}} = 1$

Second, get values of propositions after modification utilizing weights.

$Value_{P_{1}^{1}} = Mass_{P_{1}^{1}} \times Weight_{P_{1}^{1}} = 1$, $Value_{P_{2}^{2}} = \frac{2}{3}$, $Value_{P_{3}^{3}} = \frac{1}{3}$

Third, gain values after normalization.

$Value_{P_{1}^{1}}^{Final} = \frac{Value_{P_{1}^{1}}}{\sum_{i = 1 =j}^{n}Value_{P_{i}^{j}}} = \frac{1}{2}$, $Value_{P_{1}^{1}}^{Final} = \frac{1}{3}$, $Value_{P_{1}^{1}}^{Final} = \frac{1}{6}$

Forth, obtain the final value according to definition of ordinal relative belief entropy.

$INU(P_{i}^{j},P_{a}^{b}) = \sum_{j = 1}^{n} IU(P_{i}^{j},P_{a}^{b}) = 2.1087$

Then, 2.1087 is the value of measurement of degree of uncertainty of the provided ordinal frame of discernment, which takes the factor of order into consideration.

\textbf{Example 2: }Assume there are three propositions $P_{1}, P_{2}, P_{3}$. Besides, their values are listed as: $Mass_{P_{1}}= \frac{1}{4}, Mass_{P_{2}}= \frac{1}{6}, Mass_{P_{3}}= \frac{7}{12}$. In this example, all six cases are taken into account and the measurement of the ordinal frame of discernment in different situations are listed in Table \ref{2}.

\begin{table}[!htbp]
	\begin{center}
		\caption{Results of six conditions}
		\begin{tabular}{lcccccc}
			\toprule
			\multicolumn{1}{m{2cm}}{Sequence}
			&\multicolumn{3}{m{4cm}}{\centering $IU(P_{i}^{j},P_{a}^{b})$}&\multicolumn{1}{m{2.5cm}}{\centering $INU(P_{i}^{j},P_{a}^{b})$}&\multicolumn{1}{m{2.8cm}}{Deng entropy \cite{Deng2020ScienceChina}}&\multicolumn{1}{m{2.8cm}}{DP entropy \cite{articlefuzzy}}\\
			\midrule
			$P_{1}^{1}, P_{2}^{2}, P_{3}^{3}$&1.3456&0.2381&0&1.5838&1.3844&0\\
			$P_{1}^{1}, P_{2}^{3}, P_{3}^{2}$&1.1480&1.2734&0&2.4214&1.3844&0\\
			$P_{1}^{2}, P_{2}^{1}, P_{3}^{3}$&0.8170&0.4023&0&1.2194&1.3844&0\\
			$P_{1}^{2}, P_{2}^{3}, P_{3}^{1}$&0.7037&1.2170&0&1.9208&1.3844&0\\
			$P_{1}^{3}, P_{2}^{1}, P_{3}^{2}$&3.1927&0.3607&0&3.5535&1.3844&0\\
			$P_{1}^{3}, P_{2}^{2}, P_{3}^{1}$&3.2621&0.1998&0&3.4619&1.3844&0\\
			\bottomrule
		\end{tabular}
	\label{2}
	\end{center}
\end{table}

It can be easily concluded that the mass obtained by proposed ordinal relative belief entropy is changing according to the sequence of proposition. On the contrary, values of the two previously proposed entropy do not change no matter how the sequence of propositions is. It is necessary to take orders of propositions into consideration due to inner connections among propositions, which means any occurrence of certain incident has an influence on evaluations of other ones.

\textbf{Example 3: }Assume there are three propositions $P_{1}, P_{2}, P_{3}$. Besides, their values are listed as: $Mass_{P_{1}}= \frac{1}{2}, Mass_{P_{2}}= \frac{5}{12}, Mass_{P_{3}}= \frac{1}{12}$. In this example, all six cases are taken into account and the measurement of the ordinal frame of discernment in different situations are listed in Table \ref{3}.

\begin{table}[!htbp]
	\begin{center}
		\caption{Results of six conditions}
		\begin{tabular}{lcccccc}
			\toprule
			\multicolumn{1}{m{2cm}}{Sequence}
			&\multicolumn{3}{m{4cm}}{\centering $IU(P_{i}^{j},P_{a}^{b})$}&\multicolumn{1}{m{2.5cm}}{\centering $INU(P_{i}^{j},P_{a}^{b})$}&\multicolumn{1}{m{2.8cm}}{Deng entropy \cite{Deng2020ScienceChina}}&\multicolumn{1}{m{2.8cm}}{DP entropy \cite{articlefuzzy}}\\
			\midrule
			$P_{1}^{1}, P_{2}^{2}, P_{3}^{3}$&2.8174&0.8769&0&3.6943&1.3250&0\\
			$P_{1}^{1}, P_{2}^{3}, P_{3}^{2}$&3.0993&0.0909&0&3.1903&1.3250&0\\
			$P_{1}^{2}, P_{2}^{1}, P_{3}^{3}$&2.2783&1.1524&0&3.4308&1.3250&0\\
			$P_{1}^{2}, P_{2}^{3}, P_{3}^{1}$&2.5932&0.0970&0&2.6902&1.3250&0\\
			$P_{1}^{3}, P_{2}^{1}, P_{3}^{2}$&0.3431&0.9796&0&1.3228&1.3250&0\\
			$P_{1}^{3}, P_{2}^{2}, P_{3}^{1}$&0.3376&0.7111&0&1.0488&1.3250&0\\
			\bottomrule
		\end{tabular}
		\label{3}
	\end{center}
\end{table}

It can be also easily obtained that sequence plays an important role in affecting values of measurement of uncertainty. Meanwhile, previously proposed can not present these kinds of differences.

\textbf{Example 4: }Assume there are three propositions $P_{1}, P_{2}, P_{3}, P_{4_{P_{1},P_{2}}}$. And $P_{4_{P_{1},P_{2}}}$ represents the union set of $P_{1}, P_{2}$. Besides, their values are listed as: $Mass_{P_{1}}= \frac{4}{13}, Mass_{P_{2}}= \frac{3}{13}, Mass_{P_{3}}= \frac{5}{13}, Mass_{P_{4_{P_{1},P_{2}}}}= \frac{1}{13}$. In this example, all six cases are taken into account and the measurement of the ordinal frame of discernment in different situations are listed in Table \ref{4}.

\begin{table}[!htbp]
	\begin{center}
		\caption{Results of twenty-four conditions}
		\begin{tabular}{lccccccc}
			\toprule
			\multicolumn{1}{m{2cm}}{Sequence}
			&\multicolumn{4}{m{5cm}}{\centering $IU(P_{i}^{j},P_{a}^{b})$}&\multicolumn{1}{m{2.5cm}}{\centering $INU(P_{i}^{j},P_{a}^{b})$}&\multicolumn{1}{m{2.8cm}}{Deng entropy \cite{Deng2020ScienceChina}}&\multicolumn{1}{m{2.8cm}}{DP entropy \cite{articlefuzzy}}\\
			\midrule
			$P_{1}^{1}, P_{2}^{2}, P_{3}^{3}, P_{4_{P_{1},P_{2}}}^{4}$&3.0632& 1.1694&0.9688&0&5.2015&1.8262&0.0769\\[5pt]
			$P_{1}^{1}, P_{2}^{2}, P_{3_{P_{1},P_{2}}}^{4},P_{4}^{3}$&3.2833&1.2077&0.0654&0&4.5565&1.8262&0.0769\\[5pt]
			$P_{1}^{1}, P_{2}^{3}, P_{3}^{2}, P_{4_{P_{1},P_{2}}}^{4}$&2.9225&2.1779&0.4785&0&5.5790&1.8262&0.0769\\[5pt]
			$P_{1}^{1}, P_{2}^{3},  P_{3_{P_{1},P_{2}}}^{4},P_{4}^{2}$&2.9787&2.1963&0.0599&0&5.2350&1.8262&0.0769\\[5pt]
			$P_{1}^{1}, P_{2_{P_{1},P_{2}}}^{4},P_{3}^{2}, P_{4}^{3}$&3.4086&0.2130&0.2731&0&3.8948&1.3250&0.0769\\[5pt]
			$P_{1}^{1}, P_{2_{P_{1},P_{2}}}^{4},P_{3}^{3},P_{4}^{2}$&2.8173&0.2266&0.5626&0&3.6065&1.8262&0.0769\\[5pt]
			$P_{1}^{2}, P_{2}^{1},  P_{3}^{3}, P_{4_{P_{1},P_{2}}}^{4}$&2.1720&1.7218&0.9965&0&4.8904&1.8262&0.0769\\[5pt]
			$P_{1}^{2}, P_{2}^{1}, P_{3_{P_{1},P_{2}}}^{4}, P_{4}^{3} $&2.3137&1.8053&0.0676&0&4.1866&1.8262&0.0769\\[5pt]
			$P_{1}^{2}, P_{2}^{3},P_{3}^{1},   P_{4_{P_{1},P_{4}}}^{4}$&2.1179&2.2458&0.7300&0&2.8004&1.8262&0.0769\\[5pt]
			$P_{1}^{2}, P_{2}^{3},  P_{3_{P_{1},P_{2}}}^{4},P_{4}^{1},$&2.1936&2.3156&0.0642&0&4.5735&1.8262&0.0769\\[5pt]
			$P_{1}^{2}, P_{2_{P_{1},P_{2}}}^{4},P_{3}^{1},  P_{4}^{3} $&2.4691&0.2267&0.4179&0&3.1138&1.8262&0.0769\\[5pt]
			$P_{1}^{2}, P_{2_{P_{1},P_{2}}}^{4},P_{3}^{3},P_{4}^{1} $&2.3993&0.2197&0.5697&0&3.1887&1.8262&0.0769\\[5pt]
			$P_{1}^{3},P_{2}^{1}, P_{3}^{2},  P_{4_{P_{1},P_{2}}}^{4}$&3.8036&1.6023&0.46631&0&5.8723&1.8262&0.0769\\[5pt]
			$P_{1}^{3},P_{2}^{1}, P_{3_{P_{1},P_{2}}}^{4},P_{4}^{2}, $&3.8947&1.6008&0.0583&0&5.5538&1.8262&0.0769\\[5pt]
			$P_{1}^{3},P_{2}^{2},P_{3}^{1},   P_{4_{P_{1},P_{2}}}^{4}$&3.8885&1.1221&0.6916&0&5.7024&1.8262&0.0769\\[5pt]
			$P_{1}^{3},P_{2}^{2},P_{3_{P_{1},P_{2}}}^{4},P_{4}^{1} $&4.0737&1.1286&0.0605&0&5.2629&1.8262&0.0769\\[5pt]
			$P_{1}^{3},P_{2_{P_{1},P_{2}}}^{4},P_{3}^{1}, P_{4}^{2}  $&4.1261&0.1906&0.3961&0&4.7129&1.8262&0.0769\\[5pt]
			$P_{1}^{3},P_{2_{P_{1},P_{2}}}^{4},P_{3}^{2} ,P_{4}^{1} $&4.2224&0.1952&0.2617&0&4.6794&1.8262&0.0769\\[5pt]
			$P_{1_{P_{1},P_{2}}}^{4},P_{2}^{1}, P_{3}^{2}, P_{4}^{3}, $&0.4758&1.4152&0.3034&0&2.1945&1.8262&0.0769\\[5pt]
			$P_{1_{P_{1},P_{2}}}^{4},P_{2}^{1},  P_{3}^{3},P_{4}^{2}$&0.4932&0.9880&0.4501&0&1.9314&1.8262&0.0769\\[5pt]
			$P_{1_{P_{1},P_{2}}}^{4},P_{2}^{2},P_{3}^{1},  P_{4}^{3}$&0.4932&0.9880&0.4501&0&1.9314&1.8262&0.0769\\[5pt]
			$P_{1_{P_{1},P_{2}}}^{4},P_{2}^{2},P_{3}^{3},P_{4}^{1}$&0.4765&0.9630&0.6119&0&2.0514&1.8262&0.0769\\[5pt]
			$P_{1_{P_{1},P_{2}}}^{4},P_{2}^{3},P_{3}^{1}, P_{4}^{2} $&0.4324&1.7840&0.4489&0&2.6654&1.3250&0.0769\\[5pt]
			$P_{1_{P_{1},P_{2}}}^{4},P_{2}^{3},P_{3}^{2},P_{4}^{1}$&0.4450&1.8202&0.2978&0&2.5630&1.8262&0.0769\\[5pt]
			\bottomrule
		\end{tabular}
		\label{4}
	\end{center}
\end{table}

It can easily concluded that the values obtained from proposed entropy are changing from the orders of confirmations of propositions. On the contrary, when applying classic entropies to measure the condition of ordinal frame of discernment, the mass obtained does not change with the variation of orders.

\textbf{Example 5: }Assume there are three propositions $P_{1}, P_{2}, P_{3_{P_{1}, P_{2}}}$. Besides, their values after modification are listed as: $Mass_{P_{1}}= \frac{6}{17}, Mass_{P_{2}}= \frac{4}{17}, Mass_{P_{3_{P_{1}, P_{2}}}}= \frac{7}{17}$. In this example, all six cases are taken into account and the measurement of different situations are listed in Table \ref{5}.

\begin{table}[!htbp]
	\begin{center}
		\caption{Results of six conditions}
		\begin{tabular}{lcccccc}
			\toprule
			\multicolumn{1}{m{2cm}}{Sequence}
			&\multicolumn{3}{m{4cm}}{\centering $IU(P_{i}^{j},P_{a}^{b})$}&\multicolumn{1}{m{2.5cm}}{\centering $INU(P_{i}^{j},P_{a}^{b})$}&\multicolumn{1}{m{2.8cm}}{Deng entropy \cite{Deng2020ScienceChina}}&\multicolumn{1}{m{2.8cm}}{DP entropy \cite{articlefuzzy}}\\
			\midrule
			$P_{1}^{1}, P_{2}^{2}, P_{3_{P_{1}, P_{2}}}^{3}$&2.1534&0.4402&0&2.5936&1.5485&0.4117\\[5pt]
			$P_{1}^{1}, P_{2_{P_{1}, P_{2}}}^{3}, P_{3}^{2}$&2.0867&1.0039&0&3.0906&1.5485&0.4117\\[5pt]
			$P_{1}^{2}, P_{2}^{1}, P_{3_{P_{1}, P_{2}}}^{3}$&1.3065&0.7981&0&2.1046&1.5485&0.4117\\[5pt]
			$P_{1}^{2}, P_{2_{P_{1}, P_{2}}}^{3}, P_{3}^{1}$&1.2898&0.9948&0&2.2846&1.5485&0.4117\\[5pt]
			$P_{1_{P_{1}, P_{2}}}^{3}, P_{2}^{1}, P_{3}^{2}$&2.5047&0.7982&0&3.3029&1.5485&0.4117\\[5pt]
			$P_{1_{P_{1}, P_{2}}}^{3}, P_{2}^{2}, P_{3}^{1}$&2.5544&0.4353&0&2.9898&1.5485&0.4117\\[5pt]
			\bottomrule
		\end{tabular}
		\label{5}
	\end{center}
\end{table}

The proposed ordinal relative belief entropy manifest effects brought by orders of propositions in an ordinal frame of discernment. Previously proposed entropies can not indicate changes brought by orders.
\section{Conclusion}
In practical situations, everything comes in an order. Previously proposed work dose not take this factor into consideration. The main contribution of this paper is that a completely new entropy to measure degrees of uncertainty of ordinal frame of discernment is proposed. Compared with classic entropies, it is able to manifest influences brought by sequence of propositions. Without doubt, the proposed entropy provides a reasonable idea in improving results of disposing uncertainties to better to be adaptive to real life. Numerical examples also proves the efficiency and rationality of the proposed ordinal relative belief entropy.
\section*{Acknowledgment}

\bibliographystyle{elsarticle-num}
\bibliography{cite}

\begin{thebibliography}{10}
\expandafter\ifx\csname url\endcsname\relax
  \def\url#1{\texttt{#1}}\fi
\expandafter\ifx\csname urlprefix\endcsname\relax\def\urlprefix{URL }\fi
\expandafter\ifx\csname href\endcsname\relax
  \def\href#1#2{#2} \def\path#1{#1}\fi

\bibitem{Xiao2019a}
F.~Xiao, {A multiple-criteria decision-making method based on D numbers and
  belief entropy}, International Journal of Fuzzy Systems 21~(4) (2019)
  1144--1153.

\bibitem{LiubyDFMEA}
B.~liu, Y.~Deng, Risk evaluation in failure mode and effects analysis based on
  {D} numbers theory, International Journal of Computers Communications \&
  Control 14~(5) (2019) 672--691.

\bibitem{Deng2019}
X.~Deng, W.~Jiang, A total uncertainty measure for {D} numbers based on belief
  intervals, International Journal of Intelligent Systems 34~(12) (2019)
  3302--3316.

\bibitem{Xue2020entailmentIJIS}
Y.~Xue, Y.~Deng, {Entailment for Intuitionistic fuzzy sets based on generalized
  belief structures}, International Journal of Intelligent Systems 35 (2020)
  963--982.

\bibitem{Zadeh1965}
L.~A. Zadeh, Fuzzy sets, Information and control 8~(3) (1965) 338--353.

\bibitem{6608375}
R.~R. {Yager}, Pythagorean fuzzy subsets, in: 2013 Joint IFSA World Congress
  and NAFIPS Annual Meeting (IFSA/NAFIPS), 2013, pp. 57--61.
\newblock \href {https://doi.org/10.1109/IFSA-NAFIPS.2013.6608375}
  {\path{doi:10.1109/IFSA-NAFIPS.2013.6608375}}.

\bibitem{8944285}
F.~{Xiao}, A distance measure for intuitionistic fuzzy sets and its application
  to pattern classification problems, IEEE Transactions on Systems, Man, and
  Cybernetics: Systems (2019) 1--13.

\bibitem{Liu2019b}
Q.~Liu, Y.~Tian, B.~Kang, {Derive knowledge of Z-number from the perspective of
  Dempster--Shafer evidence theory}, Engineering Applications of Artificial
  Intelligence 85 (2019) 754--764.

\bibitem{Yager2014}
R.~R. Yager, On the maximum entropy negation of a probability distribution,
  IEEE Transactions on Fuzzy Systems 23~(5) (2014) 1899--1902.

\bibitem{Pan2020}
L.~Pan, Y.~Deng, {An association coefficient of belief function and its
  application in target recognition system}, International Journal of
  Intelligent Systems 35 (2020) 85--104.

\bibitem{Yager2014a}
R.~Yager, {Pythagorean Membership Grades in Multicriteria Decision Making},
  IEEE Transactions on Fuzzy Systems 22 (2014) 958--965.

\bibitem{Liu2019}
Z.~Liu, Y.~Liu, J.~Dezert, F.~Cuzzolin, Evidence combination based on credal
  belief redistribution for pattern classification, IEEE Transactions on Fuzzy
  Systems (2019) DOI: 10.1109/TFUZZ.2019.2911915.

\bibitem{Song2018}
Y.~Song, X.~Wang, W.~Wu, W.~Quan, W.~Huang, Evidence combination based on
  credibility and non-specificity, Pattern Analysis and Applications 21~(1)
  (2018) 167--180.

\bibitem{articledd}
S.~Singh, S.~Sharma, A.~Ganie, On generalized knowledge measure and generalized
  accuracy measure with applications to madm and pattern recognition,
  Computational and Applied Mathematics 39 (09 2020).
\newblock \href {https://doi.org/10.1007/s40314-020-01243-2}
  {\path{doi:10.1007/s40314-020-01243-2}}.

\bibitem{Song2019a}
Y.~Song, Q.~Fu, Y.-F. Wang, X.~Wang, Divergence-based cross entropy and
  uncertainty measures of {Atanassov's} intuitionistic fuzzy sets with their
  application in decision making, Applied Soft Computing 84 (2019) 105703.

\bibitem{Yager2009}
R.~R. Yager, Weighted maximum entropy owa aggregation with applications to
  decision making under risk, IEEE Transactions on Systems, Man, and
  Cybernetics-Part A: Systems and Humans 39~(3) (2009) 555--564.

\bibitem{Han2019}
Y.~Han, Y.~Deng, Z.~Cao, C.-T. Lin, An interval-valued {Pythagorean}
  prioritized operator based game theoretical framework with its applications
  in multicriteria group decision making, Neural Computing and Applications
  (2019) DOI: 10.1007/s00521--019--04014--1\href
  {https://doi.org/10.1007/s00521-019-04014-1}
  {\path{doi:10.1007/s00521-019-04014-1}}.

\bibitem{fei2019mcdm}
L.~Fei, Y.~Deng, Multi-criteria decision making in {Pythagorean} fuzzy
  environment, Applied Intelligence 50~(2) (2020) 537--561.

\bibitem{Fei2019}
L.~Fei, J.~Xia, Y.~Feng, L.~Liu, An {ELECTRE}-based multiple criteria decision
  making method for supplier selection using {Dempster-Shafer} theory, IEEE
  Access 7 (2019) 84701--84716.

\bibitem{articlesdsdsd}
c.-m. Own, Switching between type-2 fuzzy sets and intuitionistic fuzzy sets:
  An application in medical diagnosis, Appl. Intell. 31 (2009) 283--291.
\newblock \href {https://doi.org/10.1007/s10489-008-0126-y}
  {\path{doi:10.1007/s10489-008-0126-y}}.

\bibitem{Li2020generateTDBF}
Y.-X. Li, D.~Pelusi, Y.~Deng, Generate two dimensional belief function based on
  an improved similarity measure of trapezoidal fuzzy numbers, Computational
  and Applied Mathematics (2020) 10.1007/s40314--020--01371--9.

\bibitem{inbook}
C.~Shannon, A Mathematical Theory of Communication (1948), 2021, pp. 121--134.
\newblock \href {https://doi.org/10.7551/mitpress/12274.003.0014}
  {\path{doi:10.7551/mitpress/12274.003.0014}}.

\bibitem{articleYager}
R.~Yager, Interval valued entropies for dempster-shafer structures,
  Knowledge-Based Systems 161 (08 2018).
\newblock \href {https://doi.org/10.1016/j.knosys.2018.08.001}
  {\path{doi:10.1016/j.knosys.2018.08.001}}.

\bibitem{articleChakraborty}
D.~Chakraborty, S.~Pal, Journal pre-proofs rough video conceptualization for
  real-time event precognition with motion entropy, Information Sciences 543
  (09 2020).
\newblock \href {https://doi.org/10.1016/j.ins.2020.09.021}
  {\path{doi:10.1016/j.ins.2020.09.021}}.

\bibitem{articleTsallis}
C.~Tsallis, Nonadditive entropy: The concept and its use, European Physical
  Journal A 40 (2008) 257--266.
\newblock \href {https://doi.org/10.1140/epja/i2009-10799-0}
  {\path{doi:10.1140/epja/i2009-10799-0}}.

\bibitem{book}
G.~Shafer, {A Mathematical Theory of Evidence}, Vol.~1, 1976.
\newblock \href {https://doi.org/10.2307/j.ctv10vm1qb}
  {\path{doi:10.2307/j.ctv10vm1qb}}.

\bibitem{Dempster1967Upper}
A.~P. Dempster, {Upper and Lower Probabilities Induced by a Multi-Valued
  Mapping}, Annals of Mathematical Statistics 38~(2) (1967) 325--339.

\bibitem{Deng2020ScienceChina}
Y.~Deng, Uncertainty measure in evidence theory, SCIENCE CHINA Information
  Sciences 64 (2021) 10.1007/s11432--020--3006--9.

\bibitem{articleSunberg}
Z.~Sunberg, J.~Rogers, A belief function distance metric for orderable sets,
  Information Fusion 14 (2013) 361--373.
\newblock \href {https://doi.org/10.1016/j.inffus.2013.03.003}
  {\path{doi:10.1016/j.inffus.2013.03.003}}.

\bibitem{articleCheng}
C.~Cheng, F.~Xiao, A distance for belief functions of orderable set, Pattern
  Recognition Letters (02 2021).
\newblock \href {https://doi.org/10.1016/j.patrec.2021.02.010}
  {\path{doi:10.1016/j.patrec.2021.02.010}}.

\bibitem{yager2019generalized}
R.~R. Yager, {Generalized Dempster--Shafer structures}, IEEE Transactions on
  Fuzzy Systems 27~(3) (2019) 428--435.

\bibitem{yager1987dempster}
R.~R. Yager, {On the Dempster-Shafer framework and new combination rules},
  Information sciences 41~(2) (1987) 93--137.

\bibitem{Liu2020a}
P.~Liu, X.~Zhang, A new hesitant fuzzy linguistic approach for multiple
  attribute decision making based on {Dempster--Shafer} evidence theory,
  Applied Soft Computing 86 (2020) 105897.

\bibitem{Xiao2019b}
F.~Xiao, Generalization of {Dempster--Shafer} theory: A complex mass function,
  Applied Intelligence (2019) DOI: 10.1007/s10489--019--01617--y.

\bibitem{Xiao2020b}
F.~Xiao, {CED: A distance for complex mass functions}, IEEE Transactions on
  Neural Networks and Learning Systems (2020) DOI: 10.1109/TNNLS.2020.2984918.

\bibitem{Jiang2019Znetwork}
W.~Jiang, Y.~Cao, X.~Deng, {A novel Z-network model based on Bayesian network
  and Z-number}, IEEE Transactions on Fuzzy Systems (2019) DOI:
  10.1109/TFUZZ.2019.2918999.

\bibitem{Zhang2020a}
J.~Zhang, R.~Liu, J.~Zhang, B.~Kang, {Extension of Yager's negation of a
  probability distribution based on Tsallis entropy}, International Journal of
  Intelligent Systems 35~(1) (2020) 72--84.

\bibitem{Li2020}
S.~Li, F.~Xiao, J.~H. Abawajy, {Conflict Management of Evidence Theory Based on
  Belief Entropy and Negation}, IEEE Access 8 (2020) 37766--37774.

\bibitem{Luo2019negation}
Z.~Luo, Y.~Deng, {A matrix method of basic belief assignment's negation in
  Dempster-Shafer theory}, IEEE Transactions on Fuzzy Systems 27 (2019)
  10.1109/TFUZZ.2019.2930027.

\bibitem{Luo2020vectorIJIS}
Z.~Luo, Y.~Deng, {A vector and geometry interpretation of basic probability
  assignment in Dempster-Shafer theory}, International Journal of Intelligent
  Systems 35~(6) (2020) 944--962.

\bibitem{Gao2019b}
X.~Gao, F.~Liu, L.~Pan, Y.~Deng, S.-B. Tsai, {Uncertainty measure based on
  Tsallis entropy in evidence theory}, {International Journal of Intelligent
  Systems} {34}~({11}) ({2019}) {3105--3120}.

\bibitem{Deng2020InformationVolume}
Y.~Deng, Information volume of mass function, International Journal of
  Computers Communications \& Control 15~(6) (2020) 3983.
\newblock \href {https://doi.org/https://doi.org/10.15837/ijccc.2020.6.3983}
  {\path{doi:https://doi.org/10.15837/ijccc.2020.6.3983}}.

\bibitem{articlequantum}
X.~Gao, Y.~Deng, Quantum model of mass function, International Journal of
  Intelligent Systems 35 (11 2019).
\newblock \href {https://doi.org/10.1002/int.22208}
  {\path{doi:10.1002/int.22208}}.

\bibitem{pan2020Probability}
L.~Pan, Y.~Deng, Probability transform based on the ordered weighted averaging
  and entropy difference, International Journal of Computers Communications \&
  Control 15~(4) (2020) 3743.
\newblock \href {https://doi.org/10.15837/ijccc.2020.4.3743}
  {\path{doi:10.15837/ijccc.2020.4.3743}}.

\bibitem{xiao2019multi}
F.~Xiao, {Multi-sensor data fusion based on the belief divergence measure of
  evidences and the belief entropy}, Information Fusion 46 (2019) 23--32.

\bibitem{articlefuzzy}
D.~Dubois, H.~Prade, A note on measures of specificity for fuzzy sets,
  International Journal of General Systems 10 (02 1985).
\newblock \href {https://doi.org/10.1080/03081078508934893}
  {\path{doi:10.1080/03081078508934893}}.

\bibitem{yanhy2020entropy}
H.~Yan, Y.~Deng, An improved belief entropy in evidence theory, IEEE Access
  8~(1) (2020) 57505--57516.
\newblock \href {https://doi.org/10.1109/ACCESS.2020.2982579}
  {\path{doi:10.1109/ACCESS.2020.2982579}}.

\bibitem{Abellan2017}
J.~Abell{\'a}n, {Analyzing properties of Deng entropy in the theory of
  evidence}, Chaos, Solitons \& Fractals 95 (2017) 195--199.

\bibitem{WEI20114273}
C.-P. Wei, P.~Wang, Y.-Z. Zhang,
  \href{http://www.sciencedirect.com/science/article/pii/S0020025511002751}{Entropy,
  similarity measure of interval-valued intuitionistic fuzzy sets and their
  applications}, Information Sciences 181~(19) (2011) 4273 -- 4286.
\newblock \href {https://doi.org/https://doi.org/10.1016/j.ins.2011.06.001}
  {\path{doi:https://doi.org/10.1016/j.ins.2011.06.001}}.
\newline\urlprefix\url{http://www.sciencedirect.com/science/article/pii/S0020025511002751}

\bibitem{zhang2020extension}
J.~Zhang, R.~Liu, J.~Zhang, B.~Kang, {Extension of Yager's negation of a
  probability distribution based on Tsallis entropy}, International Journal of
  Intelligent Systems 35~(1) (2020) 72--84.

\bibitem{deng2021fuzzymembershipfunction}
J.~Deng, Y.~Deng, Information volume of fuzzy membership function,
  International Journal of Computers Communications \& Control 16~(1) (2021)
  4106.
\newblock \href {https://doi.org/https://doi.org/10.15837/ijccc.2021.1.4106}
  {\path{doi:https://doi.org/10.15837/ijccc.2021.1.4106}}.

\bibitem{article5555}
X.~Deng, Analyzing the monotonicity of belief interval based uncertainty
  measures in belief function theory, International Journal of Intelligent
  Systems 33 (03 2018).
\newblock \href {https://doi.org/10.1002/int.21999}
  {\path{doi:10.1002/int.21999}}.

\end{thebibliography}
\end{document}